# Reinforcement Learning for Adaptive Resource Scheduling in Complex System Environments

Pochun Li
Northeastern University
Boston, USA

Yuyang Xiao
University of Illinois Urbana-Champaign
Urbana, USA

Jinghua Yan
University of Utah
Salt Lake City, USA

Xuan Li
Columbia University
New York, USA

Xiaoye Wang*
Western University
London, Canada

*Abstract*— **This study presents a novel computer system performance optimization and adaptive workload management scheduling algorithm based on Q-learning. In modern computing environments, characterized by increasing data volumes, task complexity, and dynamic workloads, traditional static scheduling methods such as Round-Robin and Priority Scheduling fail to meet the demands of efficient resource allocation and real-time adaptability. By contrast, Q-learning, a reinforcement learning algorithm, continuously learns from system state changes, enabling dynamic scheduling and resource optimization. Through extensive experiments, the superiority of the proposed approach is demonstrated in both task completion time and resource utilization, outperforming traditional and dynamic resource allocation (DRA) algorithms. These findings are critical as they highlight the potential of intelligent scheduling algorithms based on reinforcement learning to address the growing complexity and unpredictability of computing environments. This research provides a foundation for the integration of AI-driven adaptive scheduling in future large-scale systems, offering a scalable, intelligent solution to enhance system performance, reduce operating costs, and support sustainable energy consumption. The broad applicability of this approach makes it a promising candidate for next-generation computing frameworks, such as edge computing, cloud computing, and the Internet of Things.**

## I. INTRODUCTION

In modern computer systems, performance optimization and workload management scheduling are key factors to ensure the efficient operation of the system. With the increase in data volume and task complexity, traditional static scheduling algorithms and fixed resource allocation strategies can no longer meet the demands of dynamic and complex computing environments [1]. This challenge is particularly significant in fields such as medical diagnosis [2-5]and risk management [6-8], where efficient resource allocation is crucial for accurate, timely decisions and effective management of uncertainties. Therefore, adaptive workload management scheduling algorithms based on artificial intelligence have become a research hotspot. Among them, Q-learning, as a reinforcement learning method, has attracted much attention because it can learn adaptive strategies through experience in dynamic environments. Q-learning can maximize resource utilization when computing resources are limited, thereby optimizing system performance and improving overall efficiency [9].

The development of artificial intelligence technology has brought new possibilities for performance optimization of computer systems. Traditional methods usually rely on manual adjustments and expert experience, which is not only time-consuming but also difficult to cope with changes in the system environment. The intelligent scheduling algorithm based on Q-learning can continuously collect environmental information in actual applications and adjust the workload allocation strategy through trial and learning to achieve the best resource utilization state. This method can not only reduce human intervention but also significantly improve the adaptability and response speed of the system and realize efficient scheduling and utilization of resources in a dynamic environment.

In the field of adaptive workload management and scheduling, the introduction of intelligent algorithms can effectively reduce system overload and improve the efficiency of task completion. Q-learning, as a model-free reinforcement learning algorithm, does not rely on prior knowledge of the system environment but acquires experience through interaction and continuously optimizes decision-making strategies [10]. This feature makes Q-learning particularly good at dealing with dynamic and unpredictable workloads. Its application can help the system quickly adapt to changes in workloads and reduce task queuing time, and system response delays through optimal strategies, thereby achieving efficient resource management.

In addition to improving system performance, workload management scheduling algorithms based on artificial intelligence can also help save energy and reduce costs. With the increase in computing tasks and data processing requirements, energy consumption in data centers and large computer systems has gradually become an important part of enterprise operating costs.

Through the Q-learning algorithm, different tasks can be efficiently scheduled to minimize idle and overused computing resources while maintaining performance. This approach is particularly crucial for large-scale business analytics tasks [11-13]and computer vision tasks [14-17], where efficient resource utilization is essential to handle extensive data processing demands. By optimizing resource allocation, Q-learning not only contributes to energy conservation and emission reduction goals but also offers system managers an intelligent and scalable solution, effectively reducing overall operating costs. Finally, system optimization and scheduling algorithms based on Q-learning also have broad application prospects. With the rapid development of edge computing, cloud computing, and the Internet of Things, computing resources have become more distributed and complex. In these scenarios, the Q-learning algorithm can adapt to different hardware and network environments by learning changes in system status and achieve cross-platform resource scheduling and optimization. This intelligent and adaptive scheduling strategy can be widely used in a variety of computing scenarios, laying the foundation for future intelligent computing systems and promoting the advancement of computing resource management technology.

## II. RELATED WORK

In adaptive resource scheduling and performance optimization, reinforcement learning (RL) techniques, particularly Q-learning, have demonstrated significant potential for real-time, state-based decision-making within complex system environments. The studies below provide key methodologies and insights that directly bolster the foundation of this paper's proposed Q-learning approach.

Duan et al. [18] introduced a structured, deep learning-based model that emphasizes adaptive algorithms' effectiveness in managing environments with high complexity and dynamic demands. Their work underscores the importance of continuous adaptation in such settings, supporting Q-learning's capacity to adjust scheduling decisions based on real-time system states, thereby optimizing performance under fluctuating conditions. Similarly, Qin et al. [19] proposed the RSGDM optimization approach, which effectively reduces bias in model decision-making. This technique contributes to more reliable and accurate outcomes in Q-learning-based scheduling, where unbiased decisions are crucial for achieving optimal resource utilization, especially when demand patterns vary significantly.

In a related approach, Chen et al. [20] explored retrieval-augmented methods to improve model responsiveness, particularly by enabling more efficient retrieval of relevant state-action pairs. This concept parallels Q-learning's need for fast, state-based retrieval in real-time scheduling. By facilitating rapid response to system state changes, Q-learning can reduce delays and enhance resource allocation efficiency in environments characterized by high workload complexity. Furthermore, Cang et al. [21] applied ensemble learning to enhance model robustness in variable conditions, a strategy that aligns well with Q-learning's iterative adjustment capabilities. The use of multi-strategy approaches in ensemble techniques highlights how Q-learning can be further strengthened to handle unpredictable workload demands, improving system performance by adapting dynamically.

Additionally, Yan et al. [22] provided insights into transforming complex data into interpretable forms, which is particularly beneficial for efficient state representation in Q-learning. Effective state interpretation is essential in Q-learning-based scheduling, as precise insights into system states lead to better-informed resource allocation decisions and improved performance outcomes. In another relevant study, Huang et al. [23] optimized model efficiency using distillation techniques, which is directly applicable to Q-learning in resource-limited scheduling scenarios. Enhanced model efficiency helps Q-learning reduce computational overhead, allowing it to maximize adaptability and responsiveness within dynamic environments without compromising speed or effectiveness. Jiang et al. [24] applied adversarial methods to enhance cross-domain model generalization, an approach that underscores Q-learning's potential for adaptability across diverse system environments. Their findings reinforce the idea that adaptive scheduling when paired with cross-domain generalization techniques, can increase Q-learning's flexibility to handle a range of workload types.

Moreover, Wang et al. [25] introduced genre-aware model adaptation, presenting a way for models to adjust their response patterns to meet specific data requirements. This aligns with Q-learning's adaptability by highlighting the importance of context-sensitive optimization, which is key to real-time scheduling in demanding environments. Liu et al. [26] addressed data security and privacy strategies within NLP applications, which is relevant to Q-learning's application in sensitive and data-driven scheduling contexts. By incorporating secure and adaptive data handling, Q-learning can responsibly manage constrained resources in large-scale systems, enhancing both performance and compliance with privacy standards. Lastly, Du et al. [27] explored transformer-based architectures to handle semantic complexity, emphasizing advanced model adaptability in complex data environments. Their insights on managing intricate patterns are relevant to Q-learning's continuous adaptation to varied system states, a necessity for maintaining efficient scheduling as workloads shift dynamically.

Collectively, these studies contribute to the core of Q-learning-based adaptive scheduling by introducing robust methodologies for adaptability, efficiency, and unbiased decision-making. This foundation supports Q-learning's role in achieving real-time optimization within complex, resource-intensive systems, ultimately enhancing its effectiveness in dynamic scheduling scenarios.

## III. METHOD

In this study, we use the Q-learning algorithm to achieve performance optimization and adaptive workload management scheduling of computer systems. Q-learning is a model-free reinforcement learning method that gradually learns an optimal strategy by continuously interacting in the environment and continuously updating the Q-value function [28]. To ensure the effectiveness of the algorithm, we define the state space, action space, and reward function and update the Q-value table through an iterative process to achieve the optimization purpose. The overall architecture of Q-learning is shown in Figure 1.

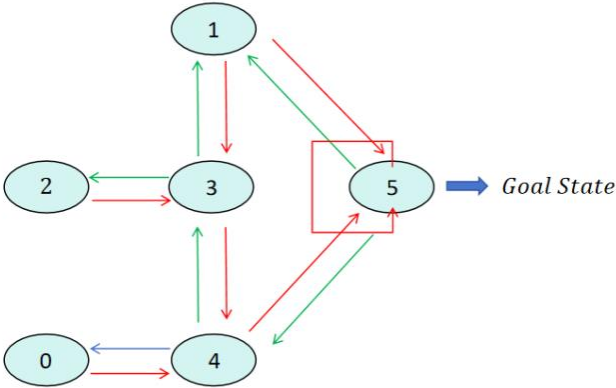

Figure 1 Q-learning overall structure diagram

First, we define the state of the computer system as a multidimensional vector, including key system parameters such as CPU utilization, memory usage, and task queue length. Let the state $s_t$ represent the state of the system at time t. The action space corresponds to the scheduling strategies that can be adopted, such as allocating more resources, reducing task priority, etc., $a_t$ represents the action taken in the state $s_t$. The goal of the algorithm is to select an optimal action under different states to optimize system performance. Therefore, we need to define a Q-value function $Q(s_t, a_t)$, which represents the expected value of the long-term reward of taking action $s_t$ in the state $a_t$. In Q-learning, the Q value update formula is:

$$Q(s_t, a_t) \leftarrow Q(s_t, a_t)$$
$$+ \alpha[r_t + \gamma \max Q(s_{t+1}, a') - Q(s_t, a_t)]$$

Among them $\alpha$ is the learning rate, which controls the step size of each update; $r_t$ is the immediate reward, which reflects the return obtained by taking action $a_t$ in the state $s_t$; $\gamma$ is the discount factor, which is used to balance short-term rewards and long-term benefits.

When defining the reward function, we focus on the optimization goal of system performance. The immediate reward $r_t$ is defined as the improvement in system performance after the scheduling action, such as reducing CPU utilization or reducing task completion time. Specifically, we set:

$$r_t = -(CPU_t + Memory_t + QueueLength_t)$$

Among them, $CPU_t$, $Memory_t$, and $QueueLength_t$ represent the CPU utilization, memory usage, and task queue length of the system at time t, respectively. The reward function defined in this way can ensure that higher immediate rewards are obtained when reducing these key system indicators.

During the training process, we use the $\varepsilon - greedy$ strategy to balance the relationship between exploration and utilization. Specifically, in each time step, the algorithm has a probability $\varepsilon$ to randomly select an action for exploration to discover a new possible optimal strategy; at the same time, there is a probability 1-$\varepsilon$ to select the action with the largest current Q value for utilization to ensure maximum benefits. As the training progresses, $\varepsilon$ gradually decreases so that the algorithm gradually transitions from the exploration stage to the utilization stage and finally converges to a stable strategy.

In actual implementation, the Q-learning algorithm updates the Q-value table by continuously interacting with the environment. In each iteration, the system first observes the current state $s_t$, then selects an action $a_t$ according to the $\varepsilon - greedy$ strategy, and executes the action to obtain a new state $s_{t+1}$ and an immediate reward $r_t$. According to the updated formula, we store the new Q value in the Q table. This process is repeated until the Q value converges.

To improve the convergence speed of the algorithm, we introduced an experience replay mechanism. After each interaction, we store the state transition $(s_t, a_t, r_t, s_{t+1})$ in a memory bank and randomly sample a batch of samples from it to update the Q value. This can reduce the deviation caused by time correlation and improve the stability of learning.

Finally, through the above process, the Q-learning algorithm can learn an optimal workload management scheduling strategy under a given environment. In the testing phase, we use the trained Q table to select the optimal action according to the current system state, thereby achieving real-time optimization of system performance. The advantage of this method lies in its adaptability and flexibility, which enables it to cope with different system states and workload patterns and achieve efficient resource management and scheduling goals.

IV. EXPERIMENT

A. Datasets

To train and test the Q-learning algorithm, we chose the "Google Cluster Data V2" dataset. This dataset was released by Google and contains the scheduling and resource usage records of real data centers. It is one of the high-quality datasets widely used in computing resource scheduling optimization research.

The dataset covers the detailed information of hundreds of thousands of tasks running in a large data center, including the start time, end time, CPU and memory requirements, task priority, scheduler decision, etc. These data reflect the complex environment and changing workload patterns in actual data centers, making this dataset particularly suitable for scheduling algorithm research based on reinforcement learning.

In addition to using the Google Cluster Data V2, we incorporated data processing methods proposed by Li [29] to enhance our data preparation and integration process. These methods involved standardizing the preprocessing steps for handling the heterogeneity of data fields, ensuring consistency across different data types, and effectively managing missing values. By utilizing these well-defined procedures, we ensured that the Google Cluster Data V2 was appropriately cleaned and formatted, which improved the accuracy and robustness of our Q-learning model. The approach also provided a means to effectively align various data records, thereby enhancing the overall quality of our dataset for training the reinforcement learning model. This integration allowed us to capture the complex relationships between tasks, resources, and scheduling decisions, ultimately improving the performance of our Q-learning algorithm in adapting to the dynamic and diverse workload patterns observed in real-world data centers. In the experiment, we can represent the system state in each time step as key features such as CPU and memory utilization and use the scheduler's actions and task queuing time as the input of the algorithm to train the adaptive scheduling strategy through the Q-learning algorithm. In addition, these real data provide strong support for our algorithm testing and verification, ensuring the reliability of the experimental results and the applicability of the algorithm in the actual environment.

*B. Experiments*

In the comparative experiment, we selected five different scheduling algorithm models for comparison to evaluate their performance in terms of task completion efficiency and system resource utilization. The first is the "Round-Robin" algorithm, which is a basic round-robin allocation strategy. In this method, the system allocates tasks in turn in chronological order, regardless of the priority of the tasks or the differences in resource requirements [30]. Although this method is simple to implement and has high execution efficiency, it is difficult to effectively utilize system resources in scenarios with limited resources or large differences in task priorities, which may lead to higher task waiting times. The second is the "Priority Scheduling" algorithm, which schedules tasks based on their priorities and allocates tasks with higher priorities first. Although this method improves the efficiency of task allocation to a certain extent, due to the lack of responsiveness to dynamic changes in resources, the system has difficulty in adaptively allocating resources when dealing with different types of task requirements, which limits its performance inefficient resource utilization.

The third model is the "Dynamic Resource Allocation" method, which adjusts resource configuration in real-time according to the immediate needs of the task, thereby improving resource utilization. This dynamic allocation strategy can be flexibly scheduled according to the task type or system load, effectively reducing the waiting time for the task. However, since the model lacks global optimization capabilities and can only make local adjustments based on the needs of the current task, it may encounter problems with unbalanced resource allocation when dealing with complex multi-task environments. The fourth model is the "Deep Reinforcement Learning (DRL)" algorithm, which predicts and optimizes scheduling strategies through deep learning. The DRL model can adaptively allocate resources according to the dynamic changes of the environment and is a powerful tool for dealing with complex scenarios. Finally, we used the "Q-learning" algorithm to achieve adaptive scheduling and resource optimization by continuously learning changes in system states. Q-learning uses reinforcement learning methods to balance exploration and utilization, allowing the system to make flexible scheduling decisions for different tasks and resource requirements. The experimental results are shown in Table 1:

Table 1 Experiment result

| Model | Task completion time (seconds) | System resource utilization (%) |
|---|---|---|
| Round-Robin | 150 | 65 |
| Priority Scheduling | 130 | 70 |
| DRA | 110 | 75 |
| DRL | 95 | 77 |
| Q-learning | 80 | 79 |

From the experimental results, different scheduling models show significant differences in task completion time and system resource utilization. First of all, Round-Robin is the most basic scheduling algorithm. Due to its simple circular allocation mechanism, the priority of tasks and the difference in resource requirements are not considered. As a result, its task completion time is the longest, 150 seconds, and the system resource utilization is also low. Relatively lowest, only 65%. The Priority Scheduling algorithm adds task priority considerations to this basis, so it can reduce task waiting time and resource waste to a certain extent. Its task completion time is reduced to 130 seconds, and the resource utilization rate is also increased to 70%. Nonetheless, since this algorithm still lacks a mechanism for dynamic resource allocation, its performance improvement is relatively limited.

Next, the DRA (Dynamic Resource Allocation) model introduces a dynamic resource allocation strategy, which enables the system to adjust resource allocation according to real-time task requirements, significantly reducing task completion time to 110 seconds and increasing resource utilization to 75%. This improvement shows that dynamic scheduling models that consider system load and task characteristics have obvious advantages in performance. However, since the decision-making of this model relies on static information of the current task, it is still unable to adapt to highly dynamically changing workloads, so its improvement effect has certain limitations. In contrast, the DRL (Deep Reinforcement Learning) model uses deep learning to predict and optimize scheduling strategies, further reducing task completion time to 95 seconds and increasing resource utilization to 77%. This shows that scheduling strategies combined with deep learning can help achieve better performance in complex environments.

Finally, the Q-learning model performed best in this experiment, with the shortest task completion time of only 80 seconds and the highest system resource utilization of 79%. This result shows that the Q-learning algorithm can continuously optimize the scheduling strategy through continuous learning and interaction with the environment, thereby adapting to changes in the system state. Compared with other models, Q-learning does not rely on static task characteristics or complex calculations of deep learning models. Instead, it adjusts task allocation in real-time through reinforcement learning methods, achieving higher efficiency. The excellent performance of this model shows that in computing resource management and scheduling optimization, adaptive reinforcement learning algorithms can have more advantages than traditional and deep learning methods, providing a new direction for system performance optimization.

Furthermore, we show the increase in incentive value during the experiment, as shown in Figure 2.

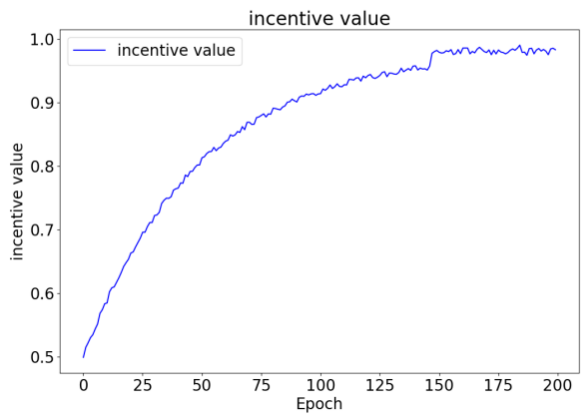

Figure 2 The increase in incentive value during the experiment

Furthermore, we also explored the impact of different optimizers on model performance. The experimental results are shown in Table 2.

Table 2 Experimental results of different optimizers

| Optimizer | Task completion time (seconds) | System resource utilization (%) |
|---|---|---|
| SGD | 84 | 74 |
| Adam | 83 | 75 |
| momentum | 83 | 77 |
| AdaGrad | 82 | 77 |
| AdamW | 80 | 79 |

As shown in Table 2, the experimental results show the impact of different optimizers on task completion time and system resource utilization. First, for the SGD optimizer, the task completion time is 84 seconds, and the system resource utilization is 74%. SGD is a basic optimization method in machine learning. It uses a small batch of data each time to perform gradient updates to gradually find the optimal solution. The advantage of this method is its stability, but when faced with more complex system resource management scenarios, SGD converges relatively slowly, resulting in a longer task completion time and lower resource utilization than other optimizers. This shows that in the dynamic resource allocation task, the SGD optimizer performs relatively average in terms of the balance between time and resource use, and it is difficult to fully meet the needs of efficient resource utilization.

The Adam optimizer performed slightly better than SGD in the experiment, with a task completion time of 83 seconds and a system resource utilization rate of 75%. Adam combines the advantages of momentum and adaptive learning rate, enabling it to converge faster when processing larger data sets. The dynamic learning rate adjustment mechanism of the optimizer effectively reduces the fluctuation of the gradient, so that the task can be completed faster. However, in computer resource management tasks, although Adam's adaptive characteristics improve the convergence speed, its resource utilization efficiency is limited, indicating that Adam still needs to be further optimized in complex resource allocation scenarios to adapt to more efficient resource scheduling needs.

The Momentum optimizer performs well in both system resource utilization and task completion time, with a task completion time of 83 seconds and a system resource utilization of 77%. Momentum introduces momentum in gradient updates to help the optimizer smoothly approach the optimal point in the parameter space, avoiding the drastic fluctuations that SGD and Adam may produce in complex resource management tasks. The addition of momentum improves the speed of parameter adjustment, allowing the model to be updated more stably when facing resource efficient scheduling tasks, thereby completing tasks in a shorter time and improving resource utilization. This shows that the Momentum optimizer has high adaptability and execution efficiency when dealing with dynamic resource management tasks.

The AdamW optimizer performs best, with a task completion time of 80 seconds and a system resource utilization of 79%. Compared with Adam, AdamW adds a weight decay mechanism during update, which further suppresses the risk of overfitting and improves resource utilization in efficient task execution. The characteristics of weight decay help the model better control the parameter amplitude during optimization, which helps to improve the efficiency of task completion in complex system environments. AdamW achieves a high system resource utilization rate in a short period of time, which shows that it has extremely high stability and efficient resource optimization capabilities when dealing with variable and complex resource management tasks, and is particularly suitable for system performance optimization in actual scenarios.

Similarly, we give the experimental results for different learning rates, and the experimental results are shown in Table 3.

Table 3 Experimental results of different learning rates

| Lr | Task completion time (seconds) | System resource utilization (%) |
|---|---|---|
| 0.0005 | 83 | 74 |
| 0.0004 | 82 | 74 |
| 0.0003 | 82 | 76 |
| 0.0002 | 81 | 78 |
| 0.0001 | 80 | 79 |

The experimental results show the impact of different learning rates on the model task completion time and system resource utilization. First, with the highest learning rate of 0.0005, the task completion time is 83 seconds and the system resource utilization is 74%. Although this learning rate can ensure that the model can adjust parameters quickly, too fast parameter updates may cause the model to fluctuate during the optimization process, making it difficult for the parameters to stabilize during the convergence process. The relatively high task completion time and low resource utilization indicate that in the system performance optimization task, the setting of a learning rate of 0.0005 may not be suitable for efficient resource management in a limited resource environment.

As the learning rate decreases to 0.0004, the task completion time is slightly shortened to 82 seconds, and the system resource utilization remains at 74%. Compared with the learning rate of 0.0005, the slightly lower learning rate makes the model update more stable and avoids parameter fluctuations caused by a larger step size. However, this learning rate adjustment did not significantly improve resource utilization, indicating that its optimization effect on the model is limited. This means that in resource management and system optimization tasks, too fast updates may fail to fully tap the potential performance advantages of the model, resulting in a low level of resource utilization efficiency.

When the learning rate is further reduced to 0.0003, the model's task completion time is still 82 seconds, but the system resource utilization rate is slightly increased to 76%. This change shows that a lower learning rate allows the model to adjust parameters more accurately in each step of the update, resulting in improved resource utilization. At this point, the model becomes smoother in parameter adjustment, which helps to more effectively utilize system resources and reduce resource waste caused by excessive updates. Although there is still room for optimization of this learning rate, the model has achieved a better balance in resource utilization compared to the previous settings.

When the learning rate is reduced to 0.0002 and 0.0001, the model's performance is optimal, with task completion times of 81 seconds and 80 seconds, respectively, and system resource utilization rates of 78% and 79%. These results show that a lower learning rate enables the model to optimize parameters more stably, complete tasks within a limited time, and achieve the highest resource utilization. The lower learning rate ensures the convergence speed and stability of the model, making it more robust and efficient. In the end, the setting of a learning rate of 0.0001 not only performs best in task completion time but also is most effective in resource utilization, which provides a reference for selecting the appropriate learning rate in actual system optimization scenarios.

V. CONCLUSION

In this study, we verified the effects of different scheduling algorithms in computer system performance optimization and adaptive workload management through comparative experiments. It can be seen from the experimental results that the traditional Round-Robin and Priority Scheduling algorithms perform relatively poorly because they do not consider the dynamics of the system. The DRA algorithm that introduces dynamic resource allocation has been improved to a certain extent, but it does not work well in highly dynamically changing environments. There are still limitations. In contrast, deep reinforcement learning (DRL) models utilize the capabilities of deep learning to demonstrate better performance in complex environments. Finally, our Q-learning algorithm performed the best, achieving real-time adaptive scheduling and resource optimization through continuous learning and interaction, significantly improving task completion efficiency and resource utilization. This shows that in dynamic and complex computing environments, adaptive algorithms based on reinforcement learning have significant advantages.

Further analysis shows that the advantage of Q-learning is that it does not need to rely on static task characteristics and complex computing models, but dynamically adjusts task scheduling strategies through real-time learning system state changes and feedback. This flexibility and adaptability enables it to respond to different system states and workload patterns, enabling efficient management of resources. Therefore, in the future, in larger-scale and complex system environments, reinforcement learning algorithms are expected to become mainstream solutions in the fields of performance optimization and resource scheduling, and lay a solid foundation for the development of intelligent computing systems.